\documentclass[sn-mathphys-num]{sn-jnl}

\usepackage{utfsym}
\usepackage{graphicx}%
\usepackage{multirow}%
\usepackage{amsmath,amssymb,amsfonts}%
\usepackage{amsthm}%
\usepackage{mathrsfs}%
\usepackage[title]{appendix}%
\usepackage{xcolor}%
\usepackage{textcomp}%
\usepackage{manyfoot}%
\usepackage{booktabs}%
\usepackage{algorithm}%
\usepackage{algorithmicx}%
\usepackage{algpseudocode}%
\usepackage{listings}%
\usepackage{subfigure}
\usepackage{caption}
\usepackage{subfloat} 

\theoremstyle{thmstyleone}%
\theoremstyle{thmstyletwo}%

\theoremstyle{thmstylethree}%

\begin{document}

\title[Article Title]{MixCut: A Data Augmentation Method for Facial Expression Recognition}


\author[1]{\fnm{Jiaxiang} \sur{Yu}}

\author[1]{\fnm{Yiyang} \sur{Liu}}

\author[1]{\fnm{Ruiyang} \sur{Fan}}

\author*[1]{\fnm{Guobing} \sur{Sun}}\email{sunguobing@hlju.edu.cn}

\affil[1]{\orgdiv{College of Electronics Engineering}, \orgname{Heilongjiang University}, \city{Harbin}, \postcode{150000}, \state{Heilongjiang}, \country{People’s
Republic of China}}


\abstract{In the facial expression recognition task, researchers always get low accuracy of expression classification due to a small amount of training samples. In order to solve this kind of problem, we proposes a new data augmentation method named MixCut. In this method, we firstly interpolate the two original training samples at the pixel level in a random ratio to generate new samples. Then, pixel removal is performed  in random square regions on the new samples to generate the final training samples. We evaluated the MixCut method on Fer2013Plus and RAF-DB. With MixCut, we achieved 85.63$\%$ accuracy in eight-label classification on Fer2013Plus and 87.88$\%$  accuracy in seven-label classification on RAF-DB, effectively improving the classification accuracy of facial expression image recognition. Meanwhile, on Fer2013Plus, MixCut achieved performance improvements of +0.59$\%$, +0.36$\%$, and +0.39$\%$ compared to the other three data augmentation methods: CutOut, Mixup, and CutMix, respectively. MixCut improves classification accuracy on RAF-DB by +0.22$\%$, +0.65$\%$, and +0.5$\%$ over these three data augmentation methods.}

\keywords{facial expression recognition,training samples, data augmentation, classification accuracy}



\maketitle

\section{Introduction}\label{sec1}

Facial expression recognition is a complex task that requires accurate description and classification of human facial expressions. Facial expression recognition based on traditional machine learning is based on extracting expression features manually and then performing expression recognition based on classification algorithms. Two traditional feature extraction methods are global feature extraction and local feature extraction. Global feature extraction is the extraction of features from the entire face image. One of the most used methods is Principal Component Analysis (PCA)\cite{dong2022denoising}. Niu et al.\cite{niu2010facial}  proposed the weighted principal component analysis method. The method adds weights to the features extracted by PCA, screening the main features further and optimizing the size of the feature matrix, thus enhancing the efficiency of facial expression recognition. Zhu et al.\cite{zhu2016face} introduced the mean principal component analysis method, optimizing the extraction of expression features and effectively enhancing the stability of facial expression classification accuracy. In addition to improving PCA itself, some researchers have improved the PCA method by combining it with other methods. Mohammadi et al.\cite{mohammadi2014pca} simulated variations in facial expressions and established a relationship model between PCA and classification dictionaries. This approach resulted in a sparse matrix representation of six types of expressions. Roweos et al.\cite{roweis2000nonlinear} used the local linear embedding method for the first time to downscale image features and reduce the number of parameters in the model. Arora M et al.\cite{arora2021autofer} hybridized PCA and particle swarm optimization algorithm to achieve high-accuracy feature vector extraction. Local feature extraction methods divide the face image into several regions and perform separate expression feature extraction for different regions. Ekman et al.\cite{ekman1978facial} proposed the classic Action Unit (AU) partition method, the Facial Action Coding System (FACS). Since then, many researchers extracted features from facial AU based on FACS for expression classification. Zhao et al.\cite{zhao2015joint} proposed multi-expression labels combined with a small region approach for feature extraction. Han et al. \cite{han2014facial} performed facial mesh transformation by facial expression to improve the AU feature extraction of the face. Hasan Z F et al. \cite{hasan2022improved} extracted the features using the Histogram of Oriented Gradient (HOG) and the Gabor method by combining the extracted features to improve facial expression detection accuracy.

In 2006, Hinton et al.\cite{hinton2006reducing} first proposed the concept of deep learning, and researchers began using Convolutional Neural Network (CNN) to extract deep features of facial expressions and classify them. Many classical CNN network models, such as AlexNet\cite{krizhevsky2017imagenet}, VGGNet\cite{simonyan2014very}, GoogLeNet\cite{szegedy2015going}, and ResNet\cite{he2016deep}, have been successfully applied to facial expression recognition. Deep learning algorithms extract image features automatically through machine training. This process eliminates the bottleneck associated with manual feature extraction, significantly improving the efficiency and accuracy of image processing. These advancements are particularly notable when compared to traditional machine learning algorithms. In the field of facial expression recognition, the research focus of deep learning techniques is mainly on improving CNN and a fusion of different CNNs. CNN is one of the most commonly used models in deep learning algorithms, and its performance improvement is crucial for the accuracy of facial expression recognition. Liu et al.\cite{liu2019real} proposed an average weighting method to reduce the error in facial expression recognition when using deep learning methods, and the results showed that the model outperformed the traditional CNN method. Kim et al.\cite{kim2021image} used training multiple convolutional neural networks and selected the one with the highest weight ratio to recognize facial expressions. Shao J et al.\cite{shao2019three} compared the effect of facial expression recognition by constructing different CNN models and concluded that shallow CNN is superior to other CNN methods. Shi et al.\cite{shi2021facial} proposed a Multibranch Cross-connection Convolutional Neural Network (MBCC-CNN), which improves the feature extraction capability of each convolutional kernel. Yu et al.\cite{yu2022co} proposed a new end-to-end Co-attentive Multitask Convolutional Neural Network (CMCNN). It comprised Channel Co-Attention Module (CCAM) and Spatial Co-Attention Module (SCAM). Many experimental results show that the method performs better than the single-task and multitask baselines. In addition, some researchers have made improvements based on the softmax loss function. Wen et al.\cite{wen2016discriminative} proposed the center loss function to achieve the minimization of intra-class variance; the central role of this function is to achieve a uniform distribution of centers within the image class. Cai et al.\cite{cai2018island} proposed an island loss function, which improves the ability to learn recognition of deep features in facial expressions. Han et al.\cite{han2022triple} proposed a triple-structured network model based on MobileNet V1, which trained a new multibranch loss function to improve expression recognition accuracy.

Deep learning algorithms have achieved great success in facial expression recognition. However, challenges still need to be solved in this task, such as the limited number of samples and the tendency of the network to overfit. Researchers widely employ data augmentation techniques in facial expression recognition to address these issues. Data augmentation technology generates more rich data samples by amplifying and transforming the original data, thus effectively improving the accuracy of facial expression recognition. Popular methods of data augmentation today include random cropping, image flipping, and random removal\cite{zhong2020random}. Researchers have evolved CNN from VGG\cite{simonyan2014very} to ResNet\cite{he2016deep} and widely used data augmentation methods. K.K. Singh et al.\cite{singh2018hide} proposed the "Hide-and-Seek" method to enhance model robustness to occlusion situations. This method generates multiple discrete hidden patches in training images, creating various occlusion combinations. Consequently, when the model encounters a hidden recognizable target during testing, it is compelled to seek other relevant content, thus improving overall robustness. S.Yun et al.\cite{yun2019cutmix} proposed that CutMix be cut and pasted randomly between training images, and the labeled data is processed accordingly. Y. Linyu et al.\cite{yan2023lmix}proposed LMix using a random mask to maintain the data distribution of the training samples and high-frequency filtering to sharpen the samples to highlight the recognition regions. T. Devries et al.\cite{devries2017improved} proposed the technique of Cutout, which involves randomly masking a fixed-size rectangular region of the input image during training. Users can combine this technique with other techniques in adversarial training. H. Zhang et al.\cite{zhang2017mixup}  proposed Mixup, which has gained widespread attention; Mixup mixes paired samples and their labels. Mixup can reduce the model's memory of mislabelling and increase robustness to adversarial instances.

\begin{figure}[H]
		\centering
		\vspace{-0.35cm}
		\subfigtopskip=2pt
		\subfigbottomskip=2pt
		\subfigcapskip=0pt
		\subfigure[Original Samples]
		{
			\includegraphics[width=0.5\linewidth]{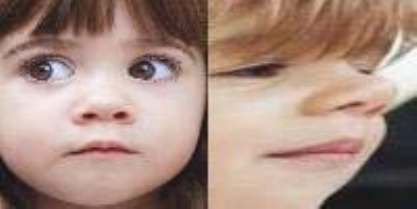}
		}
		\quad
		\subfigure[MixCut]
		{
			\includegraphics[width=0.25\linewidth]{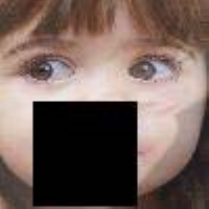}
		}

		\subfigure[Cutout]
		{
			\includegraphics[width=0.25\linewidth]{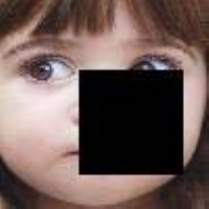}
		}
		\quad
		\subfigure[Mixup]
		{
			\includegraphics[width=0.25\linewidth]{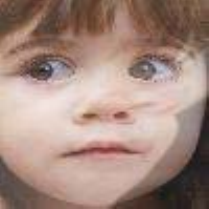}
		}
		\quad
		\subfigure[CutMix]
		{
			\includegraphics[width=0.25\linewidth]{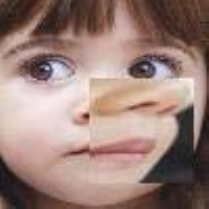}
		}
		\caption{Visual comparison of MixCut with Cutout, Mixup, and CutMix}\label{fig1}
	\end{figure}

This paper proposes a new data augmentation method combining Cutout, Mixup, and CutMix: MixCut. The MixCut method starts by interpolating two samples in a particular ratio to generate new samples and new labels in the same ratio. The new samples obtained from interpolation are then randomly removed with a portion of square pixels to obtain the new samples after final data augmentation. Figure 
\ref{fig1} shows a visual comparison of MixCut with three other data augmentation methods on RAF-DB\cite{li2017reliable}. To validate the method's effectiveness, we evaluated it on the facial expression recognition dataset Fer2013Plus\cite{barsoum2016training} and the RAF-DB dataset, with the improved VGG19 selected as the Baseline (see subsection \ref{subsec1} for details). In the Fer2013Plus dataset, compared to Baseline, MixCut's classification accuracy improved by +1.27$\%$. MixCut's classification accuracy in the RAF-DB dataset improved by +1.04$\%$ compared to the Baseline. In both datasets, the classification accuracy after data augmentation using MixCut is better than after using Cutout, Mixup, and CutMix.

\section{MixCut}\label{sec2}

MixCut is a simple data augmentation method. In simple terms, MixCut generates new samples by interpolating two original training samples randomly at the pixel level. It then removes the new samples' pixels in random square areas to create the final training samples.

Let $x\in R^{W\times H\times C}$, and $y$ denote the training samples and their labels, respectively. The goal of MixCut is to generate a new training sample $(\tilde{x},\tilde{y})$ by combining two training samples $(x_A, y_A)$ and $(x_B, y_B)$. The generated training sample $(\tilde{x},\tilde{y})$ trains a model with its original loss function. We define this operation as:
\begin{equation}
\tilde{x}=M\odot[\lambda x_A+(1-\lambda)x_B]\label{eq1}
\end{equation}
\begin{equation}
\tilde{y}=\lambda y_A+(1-\lambda)y_B\label{eq2}
\end{equation}
Where $x\in R^{W\times H\times C}$ denotes the binary mask with the exact dimensions as the training samples $x_A$ and $x_B$. $\odot$ denotes the matrix dot product. $\lambda\in[0,1]$ denotes the interpolation strength between the two training samples.

\begin{algorithm}
\caption{Pseudo-code of MixCut}\label{algo1}
\begin{algorithmic}[1]
\For{each iteration}
        \State input\_A, target\_A = get\_minibatch(dataset)\Comment{input is N×C×W×H size,  \Statex \quad \quad \quad \quad \quad \quad \quad \quad \quad \quad \quad \quad \quad \quad \quad \quad \quad \quad \quad \quad \quad \quad \quad target is is N×K size tensor.}
        \If{mode == training}
            \State input\_B, target\_B = shuffle minibatch(input\_A, target\_A)\Comment{MixCut starts.}
            \State lambda = Unif(0,1)
            \State input = lambda * input\_A + (1 - lambda) * input\_A
            \State target = lambda * target\_B + (1 - lambda) * target\_B
            \State eta = Unif(0,1)
            \State r\_x = Unif(0,W)
            \State r\_y = Unif(0,H)
            \State r\_w = Sqrt(1 - eta)
            \State r\_h = Sqrt(1 - eta)
            \State x\_1 = Round(Clip(r\_x - r\_w / 2, min=0))
            \State x\_2 = Round(Clip(r\_x + r\_w / 2, max=W))
            \State y\_1 = Round(Clip(r\_y - r\_h / 2, min=0))
            \State y\_2 = Round(Clip(r\_y + r\_h / 2, min=H))
            \State mask = ones\_like(input)
            \State mask[:,:,x\_1:x\_2,y\_1:y\_2] = 0.0
            \State input = mask * input\Comment{MixCut ends.}
        \EndIf
        \State output = model\_forward(input)
        \State loss = compute\_loss(output, target)
        \State model\_update()
\EndFor
\end{algorithmic}
\end{algorithm}

Firstly, Generating a binary mask $M^{\prime}$ with the same size as the training samples $x_A$ and $x_B$ for generating binary mask $M$, and all the elements in $M^{\prime}$ are 1. The next step involves randomly selecting the coordinates of an element in $M^{\prime}$ as the center point. We generate a square removal region $N$ around this point and complete the binary mask $M$ by setting all elements in $N$ to 0. We set the removal area ratio $\beta$, $\beta=1-\eta$, $\eta$ obey a beta distribution $Beta(1,1)$. This approach does not allow all removal areas $N$ contained in the binary mask $M^{\prime}$. The actual removal area may be smaller than the expected removal area. The interpolation strength $\lambda$ obeys a beta distribution $Beta(1,1)$. In addition, the hyperparameter $\gamma$ in MixCut controls the probability of using MixCut, and it is set to 0.5 here, indicating a 50$\%$ probability of using MixCut during the training process. Section 3.3 discusses the selection of the hyperparameters $\beta$, $\lambda$, and $\gamma$ in detail.

In each training iteration, the new MixCut samples $(\tilde{x},\tilde{y})$ are generated by randomly selecting two training samples in a small batch according to Equation \ref{eq1} and \ref{eq2}. Algorithm \ref{algo1} shows the MixCut pseudo-code.

\begin{figure}[H]
		\centering
		\vspace{-0.35cm}
		\subfigtopskip=2pt
		\subfigbottomskip=2pt
		\subfigcapskip=-10pt
		\subfigure[MixCut]
		{
			\includegraphics[width=0.73\linewidth]{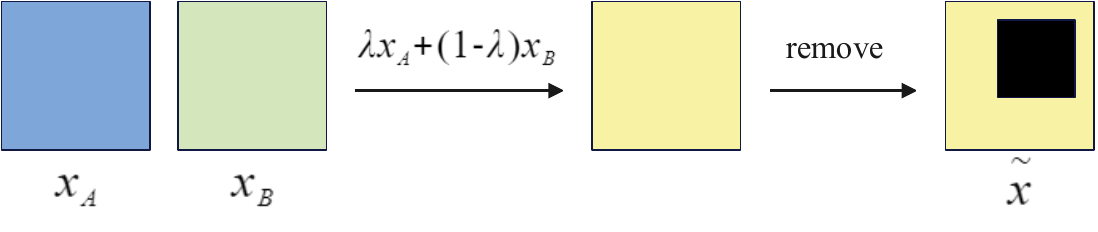}
		}

		\subfigure[Cutout]
		{
			\includegraphics[width=0.35\linewidth]{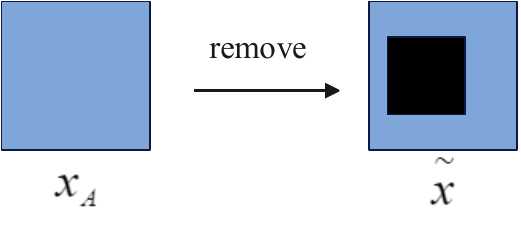}
		}
		\quad
		\subfigure[Mixup]
		{
			\includegraphics[width=0.5\linewidth]{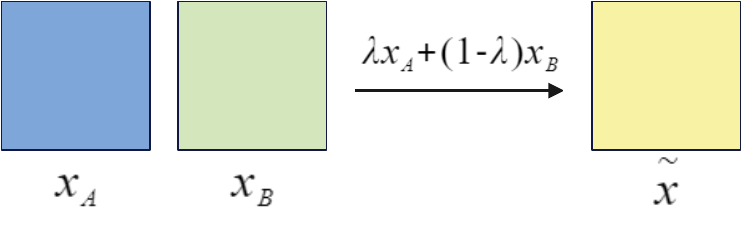}
		}

		\subfigure[CutMix]
		{
			\includegraphics[width=0.68\linewidth]{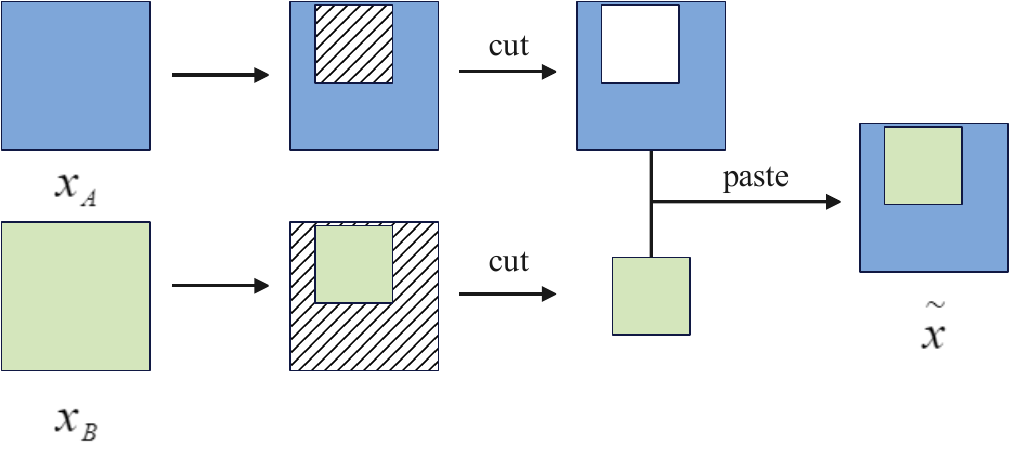}
		}
		\caption{Color represents image pixel values. The black area indicates randomly cut region. $\lambda$ represents interpolation strength, $x_A$, and $x_B$ represent original samples, and $\tilde{x}$ represents the final generated new sample}
		\label{fig2}
	\end{figure}

The Cutout\cite{devries2017improved} removes random square regions of the original sample to generate new samples. However, Cutout fixes the removal area, keeping the removal area ratio $\beta$ constant, and it does not modify the sample labels. The Mixup\cite{zhang2017mixup}  interpolates two original samples at random proportions at the pixel level to generate new samples without removing them. The new sample obtains its labels by mixing the labels of the two original samples in an interpolated ratio. The probability $\gamma$ of using Cutout and Mixup is 1. The CutMix\cite{yan2023lmix} involves cutting a random square region of one original sample and pasting it onto the cut region of another original sample to generate new samples. The new samples obtained their labels by mixing the labels of the two original samples with a cut area ratio. The probability $\gamma$ of using CutMix is 0.5. Figure \ref{fig2} illustrates the process of generating new samples from the original samples using MixCut, Cutout, Mixup, and CutMix. Table \ref{tab1} presents the main differences between MixCut, Cutout, Mixup, and CutMix.

\begin{table}[h]
\caption{Main differences between MixCut, Cutout, Mixup, and CutMix}\label{tab1}%
\begin{tabular}{@{}lllll@{}}
\toprule
 & MixCut & Cutout & Mixup & CutMix\\
\midrule
interpolation &\usym{1F5F8} & \usym{2717} & \usym{1F5F8} & \usym{1F5F8}  \\
removal &  \usym{1F5F8}  & \usym{1F5F8} & \usym{2717} & \usym{2717}  \\
mixed labels &  \usym{1F5F8}  & \usym{2717} &\usym{1F5F8} & \usym{1F5F8}  \\
original image input &  \usym{1F5F8}  & \usym{2717} & \usym{2717} & \usym{1F5F8}  \\
\botrule
\end{tabular}
\end{table}

The MixCut method is very concise and can train the network more efficiently.

\section{Experiment}\label{sec3}

To evaluate the performance of MixCut, we conducted experiments on two facial expression recognition datasets, Fer2013Plus and RAF-DB, respectively.

\subsection{Experiments on the Fer2013Plus dataset}\label{subsec1}

The Fer2013Plus dataset\cite{barsoum2016training} is a reannotation of the Fer2013 dataset\cite{goodfellow2013challenges}. In this, ten annotators annotate each image by providing their votes. In addition to the original seven expressions (anger, disgust, fear, happiness, neutral, sadness, and surprise) from FER2013, three new categories, contempt, unknown, and NF (No Face), are added. Fer2013Plus provides a more accurate representation of emotional states in static images compared to the original FER2013 labels. The dataset contains 35887 images, each with a size of 48*48 pixels, divided into three sections: 28709 training images, 3589 validation images, and 3589 test images.

Prior to the experiment, we made the following treatment of the Fer2013Plus dataset. First, we selected the label with the highest number of votes for each image as the sole label, transforming the Fer2013Plus dataset from a multi-label dataset to a single-label dataset. Second, after the above processing, we removed images labeled as unknown and NF. After this treatment, the Fer2013Plus dataset comprised 28,389 training images, 3,553 validation images, and 3,546 testing images. The exclusion of images labeled as unknown and NF was due to their lack of valuable features from which the network could learn.

\begin{figure}[h]
\centering
\includegraphics[width=0.9\textwidth]{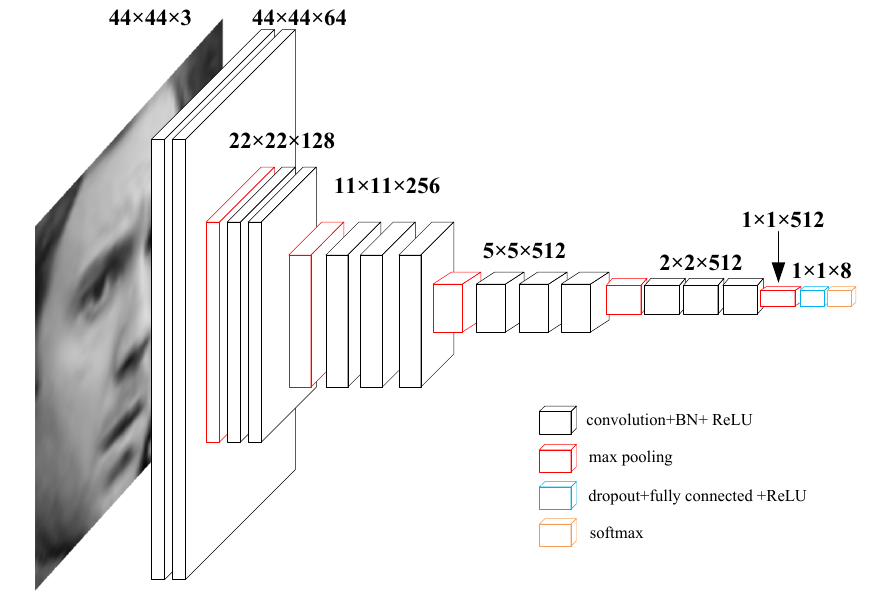}
\caption{Improved VGG19 Network Architecture Diagram}\label{fig3}
\end{figure}

We uniformly processed the obtained Fer2013Plus dataset. We randomly cropped 44x44 images during the training phase and applied random mirroring before feeding them into the training process. For the testing phase, we employed an ensemble approach to reduce outliers. We cropped and mirrored the images at the top-left, bottom-left, top-right, bottom-right, and center, resulting in 10 images from one test image. The model fed these ten images and then averaged the obtained probabilities. The classification with the highest output probability was considered the corresponding expression. This approach effectively reduced classification errors. We made several adjustments to the VGG19\cite{simonyan2014very} structure. First, we added a Batch Normalization (BN)\cite{ioffe2015batch} layer before the ReLU layers\cite{glorot2011deep}. Then, we incorporated a dropout\cite{srivastava2014dropout} strategy before the fully connected layers. Lastly, we removed multiple fully connected layers in the traditional VGG19, directly classifying them into eight categories after one fully connected layer. Figure \ref{fig3} illustrates the structure of the improved VGG19 network, which takes 44*44-sized images as input. We selected the improved VGG19 as the Baseline. The above processing includes the processing of the dataset, the preprocessing of the training data, the strategy used in the testing phase, and the selection of the improved VGG19 as the Baseline. The purpose of all of them is that the model can get better expressive power and reduce the prediction error so that the only variable of the experiment comes from different data augmentation methods.

We set the batch size to 128 in the experiment and trained for 250 epochs. Set the initial learning rate to 0.01 and decayed it every five epochs starting from the 80th epoch, multiplying it by 0.9 each time. Additionally, we utilized an SGD optimizer with a momentum of 0.9 and weight decay of 5e-4.

\begin{table}[h]
\caption{Comparison of MixCut with other data augmentation methods on Fer2013Plus}\label{tab2}%
\begin{tabular}{@{}ll@{}}
\toprule
Method & Test Acc($\%$)\\
\midrule
Baseline & 84.36  \\
Baseline+Cutout & 85.04  \\
Baseline+Mixup & 85.27  \\
Baseline+CutMix & 85.24  \\
Baseline+MixCut(ours) & 85.63  \\
\botrule
\end{tabular}
\end{table}

On the Fer2013Plus dataset, we evaluated different data augmentation methods. We set the removal ratio of CutOut to 0.25. The interpolation strength of Mixup followed a Beta distribution of $Beta(1,1)$. We set the cut ratio of CutMix to follow a Beta distribution of $Beta(1,1)$, with a 50$\%$ probability of usage. Set MixCut's hyperparameters in the same way as in section \ref{sec2}. In this experiment, the only variable was the different data augmentation methods used. We tested each method, including the Baseline, five times and then took the average result. Table \ref{tab2} shows the accuracy of the comparison of MixCut with other methods on Fer2013Plus.

MixCut achieved an accuracy of 85.63$\%$ on Fer2013Plus, which is +1.27$\%$ higher than the Baseline accuracy of 84.36$\%$. Furthermore, it achieved the highest accuracy among the four data augmentation methods, outperforming Cutout, Mixup, and CutMix by +0.59$\%$, +0.36$\%$, and +0.39$\%$, respectively.

\subsection{Experiments on the RAF-DB dataset}\label{subsec2}

RAF-DB\cite{li2017reliable}, the Real-world Affective Faces, is a large-scale facial expression dataset comprising 29,672 diverse facial images. Forty annotators are labeling images of basic or composite expressions. Additionally, each image includes annotations for five precise landmark locations, 37 automatic landmark locations, a bounding box, race, age range, and gender attributes. We selected a single-label subset of RAF-DB for this experiment, comprising 15,339 images. Each image is sized at 100*100 pixels and divided into two parts: 12,271 training images and 3,068 testing images. There are a total of 7 expression categories (anger, disgust, fear, happiness, neutral, sadness, and surprise). We did not conduct any preprocessing on the RAF-DB dataset itself. During training, images were randomly cropped to 92*92 pixels and then augmented with random mirroring before inputting into the training process. The Baseline used the same network structure as described in subsection \ref{subsec1}. Since RAF-DB has seven expression categories and the input image size is 92*92, adjustments were made to the fully connected layers accordingly. All other conditions, including the settings of data augmentation method hyperparameters, remained consistent with subsection \ref{subsec1}. We experimented with each method, including the Baseline, five times and took the average values. Table \ref{tab3} displays the accuracy of the comparison of MixCut with other methods on RAF-DB.

\begin{table}[h]
\caption{Comparison of MixCut with other data augmentation methods on RAF-DB}\label{tab3}%
\begin{tabular}{@{}ll@{}}
\toprule
Method & Test Acc($\%$)\\
\midrule
Baseline & 86.84  \\
Baseline+Cutout & 87.66  \\
Baseline+Mixup & 87.23  \\
Baseline+CutMix & 87.38  \\
Baseline+MixCut(ours) & 87.88  \\
\botrule
\end{tabular}
\end{table}

MixCut achieved a classification accuracy of 87.88$\%$ on RAF-DB, which is +1.04$\%$ higher than the baseline accuracy of 86.84$\%$. It outperformed the other three data augmentation methods, with an accuracy advantage of +0.22$\%$, +0.65$\%$, and +0.5$\%$ over Cutout, Mixup, and CutMix, respectively.

\subsection{Selection of Hyperparameters}\label{subsec3}

In this section, we discuss the settings of three hyperparameters: interpolation strength $\lambda$, removal area ratio $\beta$, and usage probability $\gamma$. The experiments in this section were all performed on the Fer2013Plus dataset after processing in subsection \ref{subsec1}. The MixCut method described in subsection \ref{subsec1} is the original MixCut, where $\lambda$ follows a Beta distribution $Beta(1,1)$; set the removal area ratio $\beta$ as 1-$\eta$, where $\eta$ follows a Beta distribution $Beta(1,1)$; and set the usage probability $\gamma$ to 50$\%$.

\subsubsection{Selection of $\lambda$}\label{subsubsec1}

For the interpolation strength $\lambda$, we set it to 0.2 and 0.4, respectively. All other conditions remain the same as the original MixCut (the discussion regarding the removal area ratio $\beta$ and the usage probability $\gamma$ is also the same). Table \ref{tab4} presents the accuracy comparison of MixCut with different interpolation strengths $\lambda$ on Fer2013Plus.

\begin{table}[h]
\caption{Comparison of accuracy of MixCut with different interpolation strengths $\lambda$ on Fer2013Plus}\label{tab4}%
\begin{tabular}{@{}ll@{}}
\toprule
Method & Test Acc($\%$)\\
\midrule
Baseline+MixCut, $\lambda$ follows $Beta(1,1)$ & 85.63  \\
Baseline+MixCut, $\lambda$=0.2 & 85.32  \\
Baseline+MixCut, $\lambda$=0.4 & 85.31  \\
\botrule
\end{tabular}
\end{table}

\subsubsection{Selection of $\beta$}\label{subsubsec2}

For the removal area ratio $\beta$, we set it to 1/4 (corresponding to a removal area edge ratio of 1/2), 9/64 (corresponding to a removal area edge ratio of 3/8), 1/16 (corresponding to a removal area edge ratio of 1/4), and 1/64 (corresponding to a removal area edge ratio of 1/8), respectively. Table \ref{tab5} compares the accuracy of MixCut with different removal area ratios $\beta$ on Fer2013Plus.

\begin{table}[h]
\caption{Comparison of accuracy of MixCut with different removal area ratios $\beta$ on Fer2013Plus}\label{tab5}%
\begin{tabular}{@{}ll@{}}
\toprule
Method & Test Acc($\%$)\\
\midrule
Baseline+MixCut, $\beta$=1-$\eta$ & 85.63  \\
Baseline+MixCut, $\beta$=1/4 & 84.90  \\
Baseline+MixCut, $\beta$=9/64 & 85.23  \\
Baseline+MixCut, $\beta$=1/16 & 85.30  \\
Baseline+MixCut, $\beta$=1/64 & 85.07  \\
\botrule
\end{tabular}
\end{table}

\subsubsection{Selection of $\gamma$}\label{subsubsec3}

For the usage probability $\gamma$,  we set it to 100$\%$. Table \ref{tab6} compares the accuracy of MixCut with different usage probabilities $\gamma$ on Fer2013Plus.

\begin{table}[h]
\caption{Presents the accuracy comparison of MixCut with different usage probabilities $\gamma$ on Fer2013Plus}\label{tab6}%
\begin{tabular}{@{}ll@{}}
\toprule
Method & Test Acc($\%$)\\
\midrule
Baseline+MixCut, $\gamma$=50$\%$ & 85.63  \\
Baseline+MixCut, $\gamma$=100$\%$ & 85.44  \\
\botrule
\end{tabular}
\end{table}

The experimental results indicate that changes in the above hyperparameters compared to the original MixCut lead to decreased MixCut's performance. The results in Tables \ref{tab4} and \ref{tab5} show that the network requires some training samples with randomness rather than setting the interpolation strength $\lambda$ and the removal area ratio $\beta$ to fixed values. Because training the network itself is inherently stochastic. The results in Table \ref{tab6} indicate that the network requires some normal or undisturbed training samples. If the training data are all subjected to noise or interference, the performance is not as good as expected.

\section{Conclusion}\label{sec4}

This paper proposes data augmentation using the MixCut method to train a convolutional neural network using the augmented samples. MixCut is easy to implement and has shown promising performance in facial expression recognition tasks. Meanwhile, we also explored the MixCut $\lambda$, $\beta$, $\gamma$ hyperparameters on the Fer2013Plus dataset and selected the best hyperparameters. On the Fer2013Plus dataset, MixCut improves the baseline classification accuracy by +1.05$\%$, reaching an accuracy of 85.63$\%$. On the RAF-DB dataset, MixCut enhances the baseline accuracy by +1.04$\%$, achieving an accuracy of 87.88$\%$. Furthermore, MixCut consistently improves classification accuracy on both datasets compared to other data augmentation methods.

MixCut performs well on the facial expression recognition task, so does it perform equally well on other classification tasks? Can it be applied to other tasks in the computer vision domain, such as detection, localization, and segmentation? Can this method be extended to semi-supervised or unsupervised learning? The high adaptability of MixCut is the focus of our future research.

\section*{Declarations}

\begin{itemize}
\item Funding: This paper was funded by Key project of Natural Science Foundation of Heilongjiang Province of China, grant number ZD2021F003.
\item Conflict of interest: The authors declare that they have no known competing financial interests or personal relationships that could have
appeared to influence the work reported in this paper.
\item Ethics approval and consent to participate: Not applicable.
\item Consent for publication: Consent.
\item Data availability: Publicly available datasets Fer2013Plus and RAF-DB.
\item Materials availability: Not applicable.
\item Code availability: Not applicable.
\item Author contribution: Conceptualization, Jiaxiang Yu; methodology, Jiaxiang Yu; software, Jiaxiang Yu; validation, Jiaxiang Yu, Yiyang Liu, Ruiyang Fan and Guobing Sun; investigation, Jiaxiang Yu; data curation, Jiaxiang Yu; writing—original draft preparation, Jiaxiang Yu; writingreview and editing, Jiaxiang Yu and Guobing Sun; visualization, Jiaxiang Yu.
\end{itemize}






\bibliography{bibliography}


\begin{thebibliography}{35}
\ifx \bisbn   \undefined \def \bisbn  #1{ISBN #1}\fi
\ifx \binits  \undefined \def \binits#1{#1}\fi
\ifx \bauthor  \undefined \def \bauthor#1{#1}\fi
\ifx \batitle  \undefined \def \batitle#1{#1}\fi
\ifx \bjtitle  \undefined \def \bjtitle#1{#1}\fi
\ifx \bvolume  \undefined \def \bvolume#1{\textbf{#1}}\fi
\ifx \byear  \undefined \def \byear#1{#1}\fi
\ifx \bissue  \undefined \def \bissue#1{#1}\fi
\ifx \bfpage  \undefined \def \bfpage#1{#1}\fi
\ifx \blpage  \undefined \def \blpage #1{#1}\fi
\ifx \burl  \undefined \def \burl#1{\textsf{#1}}\fi
\ifx \doiurl  \undefined \def \doiurl#1{\url{https://doi.org/#1}}\fi
\ifx \betal  \undefined \def \betal{\textit{et al.}}\fi
\ifx \binstitute  \undefined \def \binstitute#1{#1}\fi
\ifx \binstitutionaled  \undefined \def \binstitutionaled#1{#1}\fi
\ifx \bctitle  \undefined \def \bctitle#1{#1}\fi
\ifx \beditor  \undefined \def \beditor#1{#1}\fi
\ifx \bpublisher  \undefined \def \bpublisher#1{#1}\fi
\ifx \bbtitle  \undefined \def \bbtitle#1{#1}\fi
\ifx \bedition  \undefined \def \bedition#1{#1}\fi
\ifx \bseriesno  \undefined \def \bseriesno#1{#1}\fi
\ifx \blocation  \undefined \def \blocation#1{#1}\fi
\ifx \bsertitle  \undefined \def \bsertitle#1{#1}\fi
\ifx \bsnm \undefined \def \bsnm#1{#1}\fi
\ifx \bsuffix \undefined \def \bsuffix#1{#1}\fi
\ifx \bparticle \undefined \def \bparticle#1{#1}\fi
\ifx \barticle \undefined \def \barticle#1{#1}\fi
\bibcommenthead
\ifx \bconfdate \undefined \def \bconfdate #1{#1}\fi
\ifx \botherref \undefined \def \botherref #1{#1}\fi
\ifx \url \undefined \def \url#1{\textsf{#1}}\fi
\ifx \bchapter \undefined \def \bchapter#1{#1}\fi
\ifx \bbook \undefined \def \bbook#1{#1}\fi
\ifx \bcomment \undefined \def \bcomment#1{#1}\fi
\ifx \oauthor \undefined \def \oauthor#1{#1}\fi
\ifx \citeauthoryear \undefined \def \citeauthoryear#1{#1}\fi
\ifx \endbibitem  \undefined \def \endbibitem {}\fi
\ifx \bconflocation  \undefined \def \bconflocation#1{#1}\fi
\ifx \arxivurl  \undefined \def \arxivurl#1{\textsf{#1}}\fi
\csname PreBibitemsHook\endcsname

\bibitem[\protect\citeauthoryear{Dong et~al.}{2022}]{dong2022denoising}
\begin{barticle}
\bauthor{\bsnm{Dong}, \binits{W.}},
\bauthor{\bsnm{Wo{\'z}niak}, \binits{M.}},
\bauthor{\bsnm{Wu}, \binits{J.}},
\bauthor{\bsnm{Li}, \binits{W.}},
\bauthor{\bsnm{Bai}, \binits{Z.}}:
\batitle{Denoising aggregation of graph neural networks by using principal
  component analysis}.
\bjtitle{IEEE Transactions on Industrial Informatics}
\bvolume{19}(\bissue{3}),
\bfpage{2385}--\blpage{2394}
(\byear{2022})
\end{barticle}
\endbibitem

\bibitem[\protect\citeauthoryear{Niu and Qiu}{2010}]{niu2010facial}
\begin{bchapter}
\bauthor{\bsnm{Niu}, \binits{Z.}},
\bauthor{\bsnm{Qiu}, \binits{X.}}:
\bctitle{Facial expression recognition based on weighted principal component
  analysis and support vector machines}.
In: \bbtitle{2010 3rd International Conference on Advanced Computer Theory and
  Engineering (ICACTE)},
vol. \bseriesno{3},
pp. \bfpage{3}--\blpage{174}
(\byear{2010}).
\bcomment{IEEE}
\end{bchapter}
\endbibitem

\bibitem[\protect\citeauthoryear{Zhu et~al.}{2016}]{zhu2016face}
\begin{bchapter}
\bauthor{\bsnm{Zhu}, \binits{Y.}},
\bauthor{\bsnm{Li}, \binits{X.}},
\bauthor{\bsnm{Wu}, \binits{G.}}:
\bctitle{Face expression recognition based on equable principal component
  analysis and linear regression classification}.
In: \bbtitle{2016 3rd International Conference on Systems and Informatics
  (ICSAI)},
pp. \bfpage{876}--\blpage{880}
(\byear{2016}).
\bcomment{IEEE}
\end{bchapter}
\endbibitem

\bibitem[\protect\citeauthoryear{Mohammadi et~al.}{2014}]{mohammadi2014pca}
\begin{barticle}
\bauthor{\bsnm{Mohammadi}, \binits{M.R.}},
\bauthor{\bsnm{Fatemizadeh}, \binits{E.}},
\bauthor{\bsnm{Mahoor}, \binits{M.H.}}:
\batitle{Pca-based dictionary building for accurate facial expression
  recognition via sparse representation}.
\bjtitle{Journal of Visual Communication and Image Representation}
\bvolume{25}(\bissue{5}),
\bfpage{1082}--\blpage{1092}
(\byear{2014})
\end{barticle}
\endbibitem

\bibitem[\protect\citeauthoryear{Roweis and Saul}{2000}]{roweis2000nonlinear}
\begin{barticle}
\bauthor{\bsnm{Roweis}, \binits{S.T.}},
\bauthor{\bsnm{Saul}, \binits{L.K.}}:
\batitle{Nonlinear dimensionality reduction by locally linear embedding}.
\bjtitle{science}
\bvolume{290}(\bissue{5500}),
\bfpage{2323}--\blpage{2326}
(\byear{2000})
\end{barticle}
\endbibitem

\bibitem[\protect\citeauthoryear{Arora and Kumar}{2021}]{arora2021autofer}
\begin{barticle}
\bauthor{\bsnm{Arora}, \binits{M.}},
\bauthor{\bsnm{Kumar}, \binits{M.}}:
\batitle{Autofer: Pca and pso based automatic facial emotion recognition}.
\bjtitle{Multimedia Tools and Applications}
\bvolume{80}(\bissue{2}),
\bfpage{3039}--\blpage{3049}
(\byear{2021})
\end{barticle}
\endbibitem

\bibitem[\protect\citeauthoryear{Ekman and Friesen}{1978}]{ekman1978facial}
\begin{botherref}
\oauthor{\bsnm{Ekman}, \binits{P.}},
\oauthor{\bsnm{Friesen}, \binits{W.V.}}:
Facial action coding system.
Environmental Psychology \& Nonverbal Behavior
(1978)
\end{botherref}
\endbibitem

\bibitem[\protect\citeauthoryear{Zhao et~al.}{2015}]{zhao2015joint}
\begin{bchapter}
\bauthor{\bsnm{Zhao}, \binits{K.}},
\bauthor{\bsnm{Chu}, \binits{W.-S.}},
\bauthor{\bsnm{Torre}, \binits{F.}},
\bauthor{\bsnm{Cohn}, \binits{J.F.}},
\bauthor{\bsnm{Zhang}, \binits{H.}}:
\bctitle{Joint patch and multi-label learning for facial action unit
  detection}.
In: \bbtitle{Proceedings of the IEEE Conference on Computer Vision and Pattern
  Recognition},
pp. \bfpage{2207}--\blpage{2216}
(\byear{2015})
\end{bchapter}
\endbibitem

\bibitem[\protect\citeauthoryear{Han et~al.}{2014}]{han2014facial}
\begin{bchapter}
\bauthor{\bsnm{Han}, \binits{S.}},
\bauthor{\bsnm{Meng}, \binits{Z.}},
\bauthor{\bsnm{Liu}, \binits{P.}},
\bauthor{\bsnm{Tong}, \binits{Y.}}:
\bctitle{Facial grid transformation: A novel face registration approach for
  improving facial action unit recognition}.
In: \bbtitle{2014 IEEE International Conference on Image Processing (ICIP)},
pp. \bfpage{1415}--\blpage{1419}
(\byear{2014}).
\bcomment{IEEE}
\end{bchapter}
\endbibitem

\bibitem[\protect\citeauthoryear{Hasan}{2022}]{hasan2022improved}
\begin{barticle}
\bauthor{\bsnm{Hasan}, \binits{Z.F.}}:
\batitle{An improved facial expression recognition method using combined hog
  and gabor features}.
\bjtitle{Science Journal of University of Zakho}
\bvolume{10}(\bissue{2}),
\bfpage{54}--\blpage{59}
(\byear{2022})
\end{barticle}
\endbibitem

\bibitem[\protect\citeauthoryear{Hinton and
  Salakhutdinov}{2006}]{hinton2006reducing}
\begin{barticle}
\bauthor{\bsnm{Hinton}, \binits{G.E.}},
\bauthor{\bsnm{Salakhutdinov}, \binits{R.R.}}:
\batitle{Reducing the dimensionality of data with neural networks}.
\bjtitle{science}
\bvolume{313}(\bissue{5786}),
\bfpage{504}--\blpage{507}
(\byear{2006})
\end{barticle}
\endbibitem

\bibitem[\protect\citeauthoryear{Krizhevsky
  et~al.}{2017}]{krizhevsky2017imagenet}
\begin{barticle}
\bauthor{\bsnm{Krizhevsky}, \binits{A.}},
\bauthor{\bsnm{Sutskever}, \binits{I.}},
\bauthor{\bsnm{Hinton}, \binits{G.E.}}:
\batitle{Imagenet classification with deep convolutional neural networks}.
\bjtitle{Communications of the ACM}
\bvolume{60}(\bissue{6}),
\bfpage{84}--\blpage{90}
(\byear{2017})
\end{barticle}
\endbibitem

\bibitem[\protect\citeauthoryear{Simonyan and
  Zisserman}{2014}]{simonyan2014very}
\begin{botherref}
\oauthor{\bsnm{Simonyan}, \binits{K.}},
\oauthor{\bsnm{Zisserman}, \binits{A.}}:
Very deep convolutional networks for large-scale image recognition.
arXiv preprint arXiv:1409.1556
(2014)
\end{botherref}
\endbibitem

\bibitem[\protect\citeauthoryear{Szegedy et~al.}{2015}]{szegedy2015going}
\begin{bchapter}
\bauthor{\bsnm{Szegedy}, \binits{C.}},
\bauthor{\bsnm{Liu}, \binits{W.}},
\bauthor{\bsnm{Jia}, \binits{Y.}},
\bauthor{\bsnm{Sermanet}, \binits{P.}},
\bauthor{\bsnm{Reed}, \binits{S.}},
\bauthor{\bsnm{Anguelov}, \binits{D.}},
\bauthor{\bsnm{Erhan}, \binits{D.}},
\bauthor{\bsnm{Vanhoucke}, \binits{V.}},
\bauthor{\bsnm{Rabinovich}, \binits{A.}}:
\bctitle{Going deeper with convolutions}.
In: \bbtitle{Proceedings of the IEEE Conference on Computer Vision and Pattern
  Recognition},
pp. \bfpage{1}--\blpage{9}
(\byear{2015})
\end{bchapter}
\endbibitem

\bibitem[\protect\citeauthoryear{He et~al.}{2016}]{he2016deep}
\begin{bchapter}
\bauthor{\bsnm{He}, \binits{K.}},
\bauthor{\bsnm{Zhang}, \binits{X.}},
\bauthor{\bsnm{Ren}, \binits{S.}},
\bauthor{\bsnm{Sun}, \binits{J.}}:
\bctitle{Deep residual learning for image recognition}.
In: \bbtitle{Proceedings of the IEEE Conference on Computer Vision and Pattern
  Recognition},
pp. \bfpage{770}--\blpage{778}
(\byear{2016})
\end{bchapter}
\endbibitem

\bibitem[\protect\citeauthoryear{Liu et~al.}{2019}]{liu2019real}
\begin{bchapter}
\bauthor{\bsnm{Liu}, \binits{K.-C.}},
\bauthor{\bsnm{Hsu}, \binits{C.-C.}},
\bauthor{\bsnm{Wang}, \binits{W.-Y.}},
\bauthor{\bsnm{Chiang}, \binits{H.-H.}}:
\bctitle{Real-time facial expression recognition based on cnn}.
In: \bbtitle{2019 International Conference on System Science and Engineering
  (ICSSE)},
pp. \bfpage{120}--\blpage{123}
(\byear{2019}).
\bcomment{IEEE}
\end{bchapter}
\endbibitem

\bibitem[\protect\citeauthoryear{Kim}{2021}]{kim2021image}
\begin{barticle}
\bauthor{\bsnm{Kim}, \binits{P.W.}}:
\batitle{Image super-resolution model using an improved deep learning-based
  facial expression analysis}.
\bjtitle{Multimedia Systems}
\bvolume{27}(\bissue{4}),
\bfpage{615}--\blpage{625}
(\byear{2021})
\end{barticle}
\endbibitem

\bibitem[\protect\citeauthoryear{Shao and Qian}{2019}]{shao2019three}
\begin{barticle}
\bauthor{\bsnm{Shao}, \binits{J.}},
\bauthor{\bsnm{Qian}, \binits{Y.}}:
\batitle{Three convolutional neural network models for facial expression
  recognition in the wild}.
\bjtitle{Neurocomputing}
\bvolume{355},
\bfpage{82}--\blpage{92}
(\byear{2019})
\end{barticle}
\endbibitem

\bibitem[\protect\citeauthoryear{Shi et~al.}{2021}]{shi2021facial}
\begin{barticle}
\bauthor{\bsnm{Shi}, \binits{C.}},
\bauthor{\bsnm{Tan}, \binits{C.}},
\bauthor{\bsnm{Wang}, \binits{L.}}:
\batitle{A facial expression recognition method based on a multibranch
  cross-connection convolutional neural network}.
\bjtitle{IEEE access}
\bvolume{9},
\bfpage{39255}--\blpage{39274}
(\byear{2021})
\end{barticle}
\endbibitem

\bibitem[\protect\citeauthoryear{Yu and Xu}{2022}]{yu2022co}
\begin{barticle}
\bauthor{\bsnm{Yu}, \binits{W.}},
\bauthor{\bsnm{Xu}, \binits{H.}}:
\batitle{Co-attentive multi-task convolutional neural network for facial
  expression recognition}.
\bjtitle{Pattern Recognition}
\bvolume{123},
\bfpage{108401}
(\byear{2022})
\end{barticle}
\endbibitem

\bibitem[\protect\citeauthoryear{Wen et~al.}{2016}]{wen2016discriminative}
\begin{bchapter}
\bauthor{\bsnm{Wen}, \binits{Y.}},
\bauthor{\bsnm{Zhang}, \binits{K.}},
\bauthor{\bsnm{Li}, \binits{Z.}},
\bauthor{\bsnm{Qiao}, \binits{Y.}}:
\bctitle{A discriminative feature learning approach for deep face recognition}.
In: \bbtitle{Computer vision--ECCV 2016: 14th European Conference, Amsterdam,
  the Netherlands, October 11--14, 2016, Proceedings, Part VII 14},
pp. \bfpage{499}--\blpage{515}
(\byear{2016}).
\bcomment{Springer}
\end{bchapter}
\endbibitem

\bibitem[\protect\citeauthoryear{Cai et~al.}{2018}]{cai2018island}
\begin{bchapter}
\bauthor{\bsnm{Cai}, \binits{J.}},
\bauthor{\bsnm{Meng}, \binits{Z.}},
\bauthor{\bsnm{Khan}, \binits{A.S.}},
\bauthor{\bsnm{Li}, \binits{Z.}},
\bauthor{\bsnm{O'Reilly}, \binits{J.}},
\bauthor{\bsnm{Tong}, \binits{Y.}}:
\bctitle{Island loss for learning discriminative features in facial expression
  recognition}.
In: \bbtitle{2018 13th IEEE International Conference on Automatic Face \&
  Gesture Recognition (FG 2018)},
pp. \bfpage{302}--\blpage{309}
(\byear{2018}).
\bcomment{IEEE}
\end{bchapter}
\endbibitem

\bibitem[\protect\citeauthoryear{Han et~al.}{2022}]{han2022triple}
\begin{barticle}
\bauthor{\bsnm{Han}, \binits{B.}},
\bauthor{\bsnm{Hu}, \binits{M.}},
\bauthor{\bsnm{Wang}, \binits{X.}},
\bauthor{\bsnm{Ren}, \binits{F.}}:
\batitle{A triple-structure network model based upon mobilenet v1 and
  multi-loss function for facial expression recognition}.
\bjtitle{Symmetry}
\bvolume{14}(\bissue{10}),
\bfpage{2055}
(\byear{2022})
\end{barticle}
\endbibitem

\bibitem[\protect\citeauthoryear{Zhong et~al.}{2020}]{zhong2020random}
\begin{bchapter}
\bauthor{\bsnm{Zhong}, \binits{Z.}},
\bauthor{\bsnm{Zheng}, \binits{L.}},
\bauthor{\bsnm{Kang}, \binits{G.}},
\bauthor{\bsnm{Li}, \binits{S.}},
\bauthor{\bsnm{Yang}, \binits{Y.}}:
\bctitle{Random erasing data augmentation}.
In: \bbtitle{Proceedings of the AAAI Conference on Artificial Intelligence},
vol. \bseriesno{34},
pp. \bfpage{13001}--\blpage{13008}
(\byear{2020})
\end{bchapter}
\endbibitem

\bibitem[\protect\citeauthoryear{Singh et~al.}{2018}]{singh2018hide}
\begin{botherref}
\oauthor{\bsnm{Singh}, \binits{K.K.}},
\oauthor{\bsnm{Yu}, \binits{H.}},
\oauthor{\bsnm{Sarmasi}, \binits{A.}},
\oauthor{\bsnm{Pradeep}, \binits{G.}},
\oauthor{\bsnm{Lee}, \binits{Y.J.}}:
Hide-and-seek: A data augmentation technique for weakly-supervised localization
  and beyond.
arXiv preprint arXiv:1811.02545
(2018)
\end{botherref}
\endbibitem

\bibitem[\protect\citeauthoryear{Yun et~al.}{2019}]{yun2019cutmix}
\begin{bchapter}
\bauthor{\bsnm{Yun}, \binits{S.}},
\bauthor{\bsnm{Han}, \binits{D.}},
\bauthor{\bsnm{Oh}, \binits{S.J.}},
\bauthor{\bsnm{Chun}, \binits{S.}},
\bauthor{\bsnm{Choe}, \binits{J.}},
\bauthor{\bsnm{Yoo}, \binits{Y.}}:
\bctitle{Cutmix: Regularization strategy to train strong classifiers with
  localizable features}.
In: \bbtitle{Proceedings of the IEEE/CVF International Conference on Computer
  Vision},
pp. \bfpage{6023}--\blpage{6032}
(\byear{2019})
\end{bchapter}
\endbibitem

\bibitem[\protect\citeauthoryear{Yan et~al.}{2023}]{yan2023lmix}
\begin{barticle}
\bauthor{\bsnm{Yan}, \binits{L.}},
\bauthor{\bsnm{Zheng}, \binits{K.}},
\bauthor{\bsnm{Xia}, \binits{J.}},
\bauthor{\bsnm{Li}, \binits{K.}},
\bauthor{\bsnm{Ling}, \binits{H.}}:
\batitle{Lmix: regularization strategy for convolutional neural networks}.
\bjtitle{Signal, Image and Video Processing}
\bvolume{17}(\bissue{4}),
\bfpage{1245}--\blpage{1253}
(\byear{2023})
\end{barticle}
\endbibitem

\bibitem[\protect\citeauthoryear{DeVries and
  Taylor}{2017}]{devries2017improved}
\begin{botherref}
\oauthor{\bsnm{DeVries}, \binits{T.}},
\oauthor{\bsnm{Taylor}, \binits{G.W.}}:
Improved regularization of convolutional neural networks with cutout.
arXiv preprint arXiv:1708.04552
(2017)
\end{botherref}
\endbibitem

\bibitem[\protect\citeauthoryear{Zhang et~al.}{2017}]{zhang2017mixup}
\begin{botherref}
\oauthor{\bsnm{Zhang}, \binits{H.}},
\oauthor{\bsnm{Cisse}, \binits{M.}},
\oauthor{\bsnm{Dauphin}, \binits{Y.N.}},
\oauthor{\bsnm{Lopez-Paz}, \binits{D.}}:
mixup: Beyond empirical risk minimization.
arXiv preprint arXiv:1710.09412
(2017)
\end{botherref}
\endbibitem

\bibitem[\protect\citeauthoryear{Li et~al.}{2017}]{li2017reliable}
\begin{bchapter}
\bauthor{\bsnm{Li}, \binits{S.}},
\bauthor{\bsnm{Deng}, \binits{W.}},
\bauthor{\bsnm{Du}, \binits{J.}}:
\bctitle{Reliable crowdsourcing and deep locality-preserving learning for
  expression recognition in the wild}.
In: \bbtitle{Proceedings of the IEEE Conference on Computer Vision and Pattern
  Recognition},
pp. \bfpage{2852}--\blpage{2861}
(\byear{2017})
\end{bchapter}
\endbibitem

\bibitem[\protect\citeauthoryear{Barsoum et~al.}{2016}]{barsoum2016training}
\begin{bchapter}
\bauthor{\bsnm{Barsoum}, \binits{E.}},
\bauthor{\bsnm{Zhang}, \binits{C.}},
\bauthor{\bsnm{Ferrer}, \binits{C.C.}},
\bauthor{\bsnm{Zhang}, \binits{Z.}}:
\bctitle{Training deep networks for facial expression recognition with
  crowd-sourced label distribution}.
In: \bbtitle{Proceedings of the 18th ACM International Conference on Multimodal
  Interaction},
pp. \bfpage{279}--\blpage{283}
(\byear{2016})
\end{bchapter}
\endbibitem

\bibitem[\protect\citeauthoryear{Goodfellow
  et~al.}{2013}]{goodfellow2013challenges}
\begin{bchapter}
\bauthor{\bsnm{Goodfellow}, \binits{I.J.}},
\bauthor{\bsnm{Erhan}, \binits{D.}},
\bauthor{\bsnm{Carrier}, \binits{P.L.}},
\bauthor{\bsnm{Courville}, \binits{A.}},
\bauthor{\bsnm{Mirza}, \binits{M.}},
\bauthor{\bsnm{Hamner}, \binits{B.}},
\bauthor{\bsnm{Cukierski}, \binits{W.}},
\bauthor{\bsnm{Tang}, \binits{Y.}},
\bauthor{\bsnm{Thaler}, \binits{D.}},
\bauthor{\bsnm{Lee}, \binits{D.-H.}}, \betal:
\bctitle{Challenges in representation learning: A report on three machine
  learning contests}.
In: \bbtitle{Neural Information Processing: 20th International Conference,
  ICONIP 2013, Daegu, Korea, November 3-7, 2013. Proceedings, Part III 20},
pp. \bfpage{117}--\blpage{124}
(\byear{2013}).
\bcomment{Springer}
\end{bchapter}
\endbibitem

\bibitem[\protect\citeauthoryear{Ioffe and Szegedy}{2015}]{ioffe2015batch}
\begin{bchapter}
\bauthor{\bsnm{Ioffe}, \binits{S.}},
\bauthor{\bsnm{Szegedy}, \binits{C.}}:
\bctitle{Batch normalization: Accelerating deep network training by reducing
  internal covariate shift}.
In: \bbtitle{International Conference on Machine Learning},
pp. \bfpage{448}--\blpage{456}
(\byear{2015}).
\bcomment{pmlr}
\end{bchapter}
\endbibitem

\bibitem[\protect\citeauthoryear{Glorot et~al.}{2011}]{glorot2011deep}
\begin{bchapter}
\bauthor{\bsnm{Glorot}, \binits{X.}},
\bauthor{\bsnm{Bordes}, \binits{A.}},
\bauthor{\bsnm{Bengio}, \binits{Y.}}:
\bctitle{Deep sparse rectifier neural networks}.
In: \bbtitle{Proceedings of the Fourteenth International Conference on
  Artificial Intelligence and Statistics},
pp. \bfpage{315}--\blpage{323}
(\byear{2011}).
\bcomment{JMLR Workshop and Conference Proceedings}
\end{bchapter}
\endbibitem

\bibitem[\protect\citeauthoryear{Srivastava
  et~al.}{2014}]{srivastava2014dropout}
\begin{barticle}
\bauthor{\bsnm{Srivastava}, \binits{N.}},
\bauthor{\bsnm{Hinton}, \binits{G.}},
\bauthor{\bsnm{Krizhevsky}, \binits{A.}},
\bauthor{\bsnm{Sutskever}, \binits{I.}},
\bauthor{\bsnm{Salakhutdinov}, \binits{R.}}:
\batitle{Dropout: a simple way to prevent neural networks from overfitting}.
\bjtitle{The journal of machine learning research}
\bvolume{15}(\bissue{1}),
\bfpage{1929}--\blpage{1958}
(\byear{2014})
\end{barticle}
\endbibitem

\end{thebibliography}

\end{document}